\documentclass[pdflatex,sn-mathphys-num]{sn-jnl}

\usepackage{graphicx}%
\usepackage{multirow}%
\usepackage{amsmath,amssymb,amsfonts}%
\usepackage{amsthm}%
\usepackage{mathrsfs}%
\usepackage[title]{appendix}%
\usepackage{xcolor}%
\usepackage{textcomp}%
\usepackage{manyfoot}%
\usepackage{booktabs}%
\usepackage{algorithm}%
\usepackage{algorithmicx}%
\usepackage{algpseudocode}%
\usepackage{listings}%

\usepackage{booktabs}
\usepackage{tabularx}
\usepackage{array}
\usepackage{amsmath,amsfonts,amssymb}
\usepackage{natbib}
\usepackage{hyperref}
\usepackage{CJKutf8}
\usepackage{float}
\usepackage{afterpage}

\usepackage{tablefootnote}

\makeatletter
\renewcommand{\@@address}[2][]{%
  \g@addto@macro\auaddress{%
     \stepcounter{affn}%
     \xdef\@currentlabel{\theaffn}%
     \jmkLabel{\theaffn}%
     {\mbox{\textsuperscript{#1}#2.}\hspace{1.5em}} }%
}
\def\addressfont{\reset@font\fontsize{9bp}{11bp}\selectfont\titraggedcenter}%
\makeatother

\theoremstyle{thmstyleone}%
\theoremstyle{thmstyletwo}%

\theoremstyle{thmstylethree}%

\begin{document}

\title[ChineseBabyLM]{The First ChineseBabyLM Challenge: training data-efficient and cognitively plausible language models for Chinese}


\author*[1]{\fnm{Siyuan} \sur{Song}}\email{ss1280@princeton.edu; chinese.babylm@gmail.com}
\equalcont{These authors contributed equally to this work.}

\author[2]{\fnm{Zhiheng} \sur{Qian}}
\equalcont{These authors contributed equally to this work.}

\author[3]{\fnm{Yunhao} \sur{Zhang}}
\author[4]{\fnm{Linyang} \sur{He}}
\author[5]{\fnm{Xiaozhe} \sur{Ji}}
\author[6]{\fnm{Yingxin} \sur{Lin}}
\author[7]{\fnm{Hongao} \sur{Zhu}}
\author[2]{\fnm{Chongtian} \sur{Shao}}
\author[8]{\fnm{Chuhan} \sur{Lang}}
\author[2]{\fnm{Luan} \sur{Li}}
\author[2]{\fnm{Rui} \sur{Wang}}
\author[5]{\fnm{Renfen} \sur{Hu}}
\author[8]{\fnm{Shaonan} \sur{Wang}}
\author*[8]{\fnm{Hai} \sur{Hu}}\email{hai.hu@polyu.edu.hk}

\affil[1]{\orgname{Princeton University}}
\affil[2]{\orgname{Shanghai Jiao Tong University}}
\affil[3]{\orgname{Chinese Academy of Sciences}}
\affil[4]{\orgname{Columbia University}}
\affil[5]{\orgname{Beijing Normal University}}
\affil[6]{\orgname{Tsinghua University}}
\affil[7]{\orgname{University of California San Diego}}
\affil[8]{\orgname{The Hong Kong Polytechnic University}}


\abstract{
This paper presents the first ChineseBabyLM Challenge, organized as part of NLPCC 2026. The challenge asked
participants to train language models from scratch using no more than 102M Chinese words. The models were evaluated on three tracks: natural language understanding, cognitive alignment,
and Hanzi knowledge. There were no restrictions on tokenizers, model architectures, or the number of
training epochs. Eighteen teams submitted 28 distinct models, generating 74 result files. The
overall-winning team used a DeBERTa-v2 architecture and introduced an auxiliary pinyin-prediction objective
during pretraining. Several submissions also explored curriculum-learning strategies and architectural
innovations. Overall, the challenge provides a benchmark for advancing data-efficient and cognitively
plausible approaches to Chinese language modeling.
}



\maketitle

\section{Introduction}

Large language models have achieved strong performance across many language
understanding tasks, but their success has typically depended on pretraining
with very large corpora and substantial compute.
This data-hungry paradigm contrasts sharply with human language acquisition:
children acquire robust linguistic competence from comparatively limited,
naturalistic input.  The BabyLM Challenge was
created to make this contrast experimentally useful by asking participants to
train language models under developmentally motivated data budgets, encouraging
research on sample-efficient pretraining, cognitively plausible data, and
evaluation protocols that go beyond raw scale~\citep{first-babylm}.

The original BabyLM shared tasks focused primarily on English.  Recent
multilingual efforts such as BabyBabelLM extend the same motivation to a wider
set of languages, showing that data-efficient language modeling should be
studied under typologically and orthographically diverse
conditions~\citep{babybabellm}.  Chinese is an especially important test case.  It lacks
explicit word boundaries, makes extensive use of compounding, permits flexible
syntactic configurations, and uses a logographic writing system in which
characters carry visual, structural, and phonological regularities.  These
properties make Chinese language learning a poor fit for evaluation protocols
designed only around alphabetic, whitespace-delimited languages, and they raise
distinct questions about tokenization, character-level representation, and the
amount of data needed to acquire linguistic generalizations.

The ChineseBabyLM Challenge is a shared task for studying data-efficient and
cognitively plausible language modeling for Chinese.  Participating systems must
be pretrained from randomly initialized weights, either on the official
102-million-word corpus derived from the Chinese portion of
BabyBabelLM~\citep{babybabellm} or on a self-compiled corpus within the same
word budget.  This setup preserves the central BabyLM constraint while allowing
participants to explore different model architectures, tokenizers, data
selection strategies, and training methods for Chinese.

The challenge evaluates models along three complementary dimensions.  The
Natural Language Understanding track combines zero-shot minimal-pair evaluation,
including ZhoBLiMP~\citep{2026zhoblimp}, with supervised Chinese understanding
tasks from CLUE~\citep{xu-etal-2020-clue}.  The Hanzi track directly tests whether models acquire structural and
phonological knowledge about Chinese characters from limited input.  The
Cognitive Modeling track uses Chinese fMRI benchmarks from
MulCogBench~\citep{zhang2025mulcogbench} to measure how well model
representations predict human neural responses during word- and discourse-level
language comprehension.  Together, these tracks are intended to reward models
that are not only accurate on downstream tasks, but also sensitive to
Chinese-specific linguistic structure and aligned with cognitively motivated
evaluation criteria.

This paper describes the design of the first ChineseBabyLM Challenge, including
the training rules, evaluation tracks, task inventory, baseline systems, performance of the 
baseline models and models submitted by participating teams.  By providing a standardized corpus, an open evaluation
pipeline, hidden final-evaluation tasks, and reproducible baselines, the shared
task aims to support a research community around Chinese data-efficient language
modeling and to contribute a language-specific counterpart to the broader
BabyLM program.

\section{Evaluation Tasks}
\label{sec:evaluation_tasks}

The ChineseBabyLM Challenge evaluates data-efficient Chinese language models along
three complementary dimensions of linguistic competence: natural language
understanding, character-level knowledge, and cognitive plausibility.  The final
evaluation consists of 14 tasks grouped into three tracks: eight Natural Language
Understanding (NLU) tasks, four Hanzi tasks, and two Cognitive Modeling tasks.

\paragraph{Evaluation formats.}
The tasks use three evaluation formats.  First, the zero-shot minimal-pair tasks
evaluate whether a model assigns a higher score to a well-formed or semantically
plausible sentence than to a minimally different ill-formed or implausible one;
the task score is the accuracy over all pairs.  For the Hanzi tasks, if the
target character or character components are not recognized by the tokenizer
(UNK), then the model is considered to fail on that minimal pair.

Second, the CLUE-style NLU tasks are evaluated by task-specific fine-tuning for
sequence classification or multiple-choice classification.

Third, the cognitive
modeling tasks evaluate whether model representations can predict human fMRI
responses: model hidden states are used as features in an encoding model, and the
predicted neural responses are compared with observed fMRI responses using
correlation-based metrics.

\begin{table*}[t]
\centering
\footnotesize
\begin{tabularx}{\textwidth}{lllXX}
\toprule
Track & Task & Split & Size in final files & Target ability \\
\midrule
NLU &
\textsc{ZhoBLiMP} &
Open &
35,400 minimal pairs, from 118 paradigms &
Zero-shot grammatical and semantic judgments \\

NLU &
\textsc{XCOMPS-ZH} &
Hidden &
14,368 minimal pairs &
Zero-shot concept--property compatibility judgments \\

NLU &
\textsc{AFQMC} &
Open &
34,334 train / 4,316 dev records &
Sentence-pair semantic similarity \\

NLU &
\textsc{OCNLI} &
Open &
50,437 train / 2,950 dev records &
Natural language inference \\

NLU &
\textsc{TNEWS} &
Open &
53,360 train / 10,000 dev records &
Short-text news classification from news titles \\

NLU &
\textsc{CLUEWSC2020} &
Open &
1,244 train / 304 dev records &
Chinese Winograd-style coreference resolution \\

NLU &
\textsc{C3} &
Hidden &
11,869 train / 3,816 dev records &
Multiple-choice Chinese machine reading comprehension \\

NLU &
\textsc{diagnostic\_nli} &
Hidden &
50,437 train / 2,122 dev records &
Diagnostic natural language inference \\

\midrule
Hanzi &
\texttt{hanzi\_structure} &
Open &
2,000 minimal pairs &
Character-level structural knowledge \\

Hanzi &
\texttt{hanzi\_structure\_hidden} &
Hidden &
2,000 minimal pairs &
Held-out character-structure minimal pairs  \\

Hanzi &
\texttt{hanzi\_pinyin} &
Open &
2,000 minimal pairs &
Character-level phonological knowledge \\

Hanzi &
\texttt{hanzi\_pinyin\_hidden} &
Hidden &
2,000 minimal pairs &
Held-out character-phonology minimal pairs \\

\midrule
Cog &
\texttt{word\_fmri} &
Open/Hidden &
672 word stimuli in the source fMRI dataset &
Brain-aligned word-level evaluation: fit model representations to fMRI
responses \\

Cog &
\texttt{fmri} &
Open/Hidden &
60 story stimuli, 52,269 words in the source fMRI dataset &
Brain-aligned discourse-level evaluation: fit model representations to fMRI
responses \\
\bottomrule
\end{tabularx}
\caption{Task inventory for the final ChineseBabyLM evaluation.  Sizes for NLU
and Hanzi tasks are from the final evaluation files; sizes for the Cog tasks
describe the corresponding Chinese fMRI stimuli from MulCogBench.}
\label{tab:chinesebabylm_tasks}
\end{table*}

\subsection{Natural Language Understanding Track}
\label{subsec:nlu_tasks}

The NLU track measures syntactic and semantic understanding in Chinese.  It
contains two zero-shot minimal-pair tasks and six supervised fine-tuning tasks,
all drawn from established Chinese benchmarks;
Table~\ref{tab:chinesebabylm_tasks} summarizes the inventory, splits, and sizes.

The zero-shot tasks test whether a pretrained model prefers the acceptable member
of a minimal pair without task-specific training.  The open task is
ZhoBLiMP~\citep{2026zhoblimp}, whose 118 paradigms (35,400 pairs in total) cover
a broad range of Chinese syntactic and semantic phenomena.  The hidden task is
XCOMPS-ZH~\citep{he-etal-2025-xcomps}, 14,368 pairs targeting concept--property
compatibility, which complements the primarily morphosyntactic focus of ZhoBLiMP
with commonsense-like semantic judgments.

The six supervised tasks are evaluated by task-specific fine-tuning.  The open
tasks are AFQMC (sentence-pair semantic matching), TNEWS (short-text news
classification), and CLUEWSC2020 (Winograd-style coreference) from
CLUE~\citep{xu-etal-2020-clue}, together with OCNLI (natural language
inference)~\citep{hu-etal-2020-ocnli}.  The hidden tasks are C3, a
multiple-choice reading comprehension benchmark~\citep{c3}, and a diagnostic NLI
task~\citep{diagnostic-nli}; both provide held-out evaluation beyond the open
development tasks.

\subsection{Cognitive Modeling Track}
\label{subsec:cog_tasks}

The Cognitive Modeling track evaluates whether model representations align with human brain responses during Chinese language comprehension. The two tasks are derived from the Chinese fMRI components of MulCogBench \citep{zhang2025mulcogbench}. In both tasks, the
evaluation pipeline extracts model representations and fits an encoding model from those representations to fMRI responses. Better prediction performance indicates that the semantic information encoded in the model’s representations is more similar to that encoded in the brain’s neural representations elicited by the corresponding linguistic stimuli \citep{zhang2024navigating}.

The \texttt{word\_fmri} task is a word-level brain-alignment benchmark based on a Chinese word fMRI dataset introduced by \citet{wang2022fmri}. The dataset contains neural responses from 11 participants as they read 672 isolated Chinese words.  The task evaluates whether representations of individual words encode information predictive of the spatial patterns of fMRI responses elicited during word comprehension. This task therefore probes relatively local lexical and semantic representations.

The \texttt{discourse\_fmri} task is a discourse-level brain-alignment benchmark based on a Chinese discourse fMRI dataset introduced by \citet{wang2022synchronized}. The dataset contains neural responses from 12 native Chinese speakers as they listened to 60 stories comprising 52,269 words in total. Compared
with the word-level task, this task evaluates representations in a more natural
and context-rich comprehension setting.  It tests whether models encode
information useful for predicting neural activity during extended spoken
discourse, where lexical, syntactic, semantic, and discourse-contextual
information must be integrated over time.

\subsection{Hanzi Track}
\label{subsec:hanzi_tasks}

The Hanzi track evaluates knowledge of properties specific to the Chinese
writing system, namely Hanzi structure and Hanzi pinyin (see Table~\ref{tab:hanzi_examples}).  Unlike the CLUE tasks, which focus on sentence- or passage-level
understanding, the Hanzi tasks test whether a model has learned character-level
regularities from limited pretraining data.  All four Hanzi tasks are evaluated
as zero-shot minimal-pair tasks.  Each file contains 2,000 minimal pairs.

For the two Hanzi structure tasks,
target characters are drawn from the Level-1 list of the
\textit{General Standard Chinese Character Table}~\citep{gsc2013}.
Character structures and component decompositions follow the Chinese
Character Dataset of \citet{wu2025visual}.
Phonetic-role and IDS-based annotations use \texttt{ids-analysis.txt}
from CJKVI-IDS~\citep{kawabata2018cjkvi}.
Target character components must appear at least 50 times in the
official pretraining corpus.

For Hanzi pinyin tasks, target characters are restricted to single-pronunciation
(monophonic) characters drawn from the 3,500 most frequent Chinese characters, so
that every item has an unambiguous reading.  Pinyin readings are obtained with
\texttt{pypinyin} (manually corrected where mislabeled) and mapped to IPA with
\texttt{Dragonmapper}, and each syllable is decomposed into three independent
dimensions: the \emph{initial} (onset consonant), the \emph{final} (rime), and the
lexical \emph{tone} (1--4).  All pinyin items take the form of homophone judgments:
each minimal pair contrasts a true statement that two characters share an identical
reading with a minimally different false statement in which the second character is
replaced by one that differs in exactly one of the three dimensions, yielding the
\texttt{initial\_diff}, \texttt{rhyme\_diff}, and \texttt{tone\_diff} conditions.
There is also a \texttt{disjoint} condition, where the two characters have completely different initials, rhymes and tones.
To ensure that the judgment reflects phonological knowledge rather than lexical frequency,
the two characters in each pair are matched on corpus frequency via Zipf scores.



\begin{table}[t]
\centering
\footnotesize
\begin{CJK*}{UTF8}{gbsn}
\begin{tabular}{ll>{\raggedright\arraybackslash}p{0.34\textwidth}>{\raggedright\arraybackslash}p{0.34\textwidth}}
\toprule
Task & Subtype & Good example & Bad example \\
\midrule
Structure & whole-to-component &
阁字外围的部分是门。 &
阁字外围的部分是包。 \\
Structure & component-to-whole &
外围是门，内部是圭，这个字是闺。 &
外围是门，内部是圭，这个字是盔。 \\
\midrule
Pinyin & \texttt{initial\_diff} &
“交”和“郊”的声母、韵母和声调完全相同。(jiao1, jiao1) &
“交”和“销”的声母、韵母和声调完全相同。(jiao1, xiao1) \\
Pinyin & \texttt{rhyme\_diff} &
“机”和“鸡”的声母、韵母和声调完全相同。(ji1, ji1) &
“机”和“兢”的声母、韵母和声调完全相同。(ji1, jing1) \\
Pinyin & \texttt{tone\_diff} &
“欲”和“芋”的声母、韵母和声调完全相同。(yu4, yu4) &
“欲”和“于”的声母、韵母和声调完全相同。(yu4, yu2) \\
Pinyin & \texttt{disjoint} &
“纠”和“鸠”的声母、韵母和声调完全相同。(jiu1, jiu1) &
“纠”和“缠”的声母、韵母和声调完全相同。(jiu1, chan2) \\
\bottomrule
\end{tabular}
\end{CJK*}
\caption{Representative minimal-pair examples in the Hanzi track.  Structure items
test character composition; pinyin items are homophone judgments built on the
template ``X and Y have identical initial, final, and tone''.  The good example is
a true statement (Y is a homophone of X); the bad example replaces Y with a
distractor that differs in exactly one dimension (\texttt{initial\_diff},
\texttt{rhyme\_diff}, \texttt{tone\_diff}) or in all three (\texttt{disjoint}).
Tone-numbered pinyin is shown in parentheses.}
\label{tab:hanzi_examples}
\end{table}



\section{Rules for model training}
All participating groups are required to pretrain their models from scratch (that is, randomly initialized weights). Loading pretrained checkpoints or distilling from existing large models is not allowed.

There are two options for pretraining data.
Option 1 is to use the official corpus released by the organizers, a 102M-word training corpus sampled from the BabyBabelLM project~\citep{babybabellm}\footnote{\url{https://huggingface.co/datasets/chinese-babylm-org/babylm-zho-100M}}.
Option 2 is to use a self-compiled pretraining corpus, which should contain at most 102M words counted via Jieba 0.42.1 in default mode~\footnote{\url{https://github.com/fxsjy/jieba}}.
There is no restriction on the number of training epochs.


We do not have any special requirements on the tokenizer or model architecture.
Note that the two Hanzi tasks are designed specifically to encourage innovations on tokenizer design so that the models encode natively the phonological and orthographical information of Chinese characters.

\section{Baseline models and results}
\label{sec:baseline_models}
We adopted six models as baselines for the first ChineseBabyLM challenge. In addition to four models pretrained from scratch, we evaluated Qwen3-0.6B\footnote{\url{https://huggingface.co/Qwen/Qwen3-0.6B}} and BERT-base-Chinese\footnote{\url{https://huggingface.co/google-bert/bert-base-chinese}} as two established open-source models trained on large-scale corpora. The implementation details of the baselines are as follows. 

For tokenizers, we randomly sampled roughly 0.8B characters from CCI3-HQ~\citep{cci3-hq}, a high-quality Chinese pretraining corpus, and trained a byte-level BPE tokenizer with a $50{,}000$ vocabulary size and a minimum frequency threshold of $2$. The tokenizer was used across all the baseline models we trained from scratch.

In terms of model architecture, 
we trained four architecture-diverse baseline models: an encoder-decoder Transformer following T5~\citep{t5}, a bidirectional encoder following BERT~\citep{bert}, a selective state-space model following Mamba~\citep{mamba}, and a decoder-only Transformer following Pythia~\citep{pythia}. 
The decoder-only Pythia and Mamba baselines are optimized with causal language modeling, while the BERT baseline uses masked language modeling with a masking probability of 15\%. The T5 baseline is trained as a text-to-text continuation model: each training sequence is split into a source prefix and target continuation, and the model is optimized to generate the continuation from the prefix.
To make the comparison computationally controlled, all architectures are scaled down to approximately 14M parameters and are trained from scratch on the official ChineseBabyLM training corpus. 

The baseline models were trained for 20 epochs using the Adam optimizer, with a learning rate of $5\times10^{-4}$, a batch size of 32, and a 1\% warmup ratio. The context length was set to 1024. The 3-, 10-, and 20-epoch checkpoints were released to the candidates.

\begin{table}[htbp]
\centering
\begin{tabularx}{\textwidth}{lXXX}
\toprule
Architecture & NLU Track & Hanzi Track & Cog Track \\
\midrule
T5     & 54.17 & 50.71 & 31.63 \\
BERT   & \underline{57.65} & 51.86 & \underline{33.38} \\
Mamba  & 56.89 & \underline{52.49} & 28.28 \\
Pythia & 55.88 & 51.73 & 29.28 \\
\hline
Qwen3-0.6B       & 57.82 & \textbf{61.24} & \textbf{34.85} \\
BERT-base-Chinese & \textbf{66.00} & 57.13 & 33.99 \\
\bottomrule
\end{tabularx}
\caption{Average performance of the baseline models by track.
Bold indicates the best result among all evaluated models, while
underlining indicates the best result among the four 14M models
pretrained from scratch.}
\label{tab:model_performance_avg}
\end{table}

Table~\ref{tab:model_performance_avg} presents the average performance of all baseline models across three tracks. Among the four 14M models pretrained from scratch, BERT achieves the highest average scores on the NLU and Cognitive Modeling tracks, whereas Mamba performs best on the Hanzi track. Among the two external reference models, BERT-base-Chinese obtains the highest NLU score, while Qwen3-0.6B leads on the Hanzi and Cognitive Modeling tracks. These two models do not satisfy the ChineseBabyLM data constraint and are only reported as external reference points.

\section{Shared Task}
\label{sec:shared_task}

\subsection{Overview}

The final stage of the first ChineseBabyLM Challenge received submissions from
18 teams.  In total, the teams submitted 28 distinct models, with several teams
entering more than one model or rerunning evaluations during the submission
period.  The evaluation records contain 74 result files across these initial
submissions and reruns.  For the official team-level ranking, one system from
each team was retained, resulting in 18 systems on the final leaderboard.

For every ranked system, task scores are reported on a 0--100 scale.  Each track
score is the unweighted mean of its constituent tasks, and the overall score is
the macro-average of the NLU, Cognitive Modeling, and Hanzi track scores.
Missing task results are assigned a score of zero.  This track-level
macro-average prevents tracks with more constituent benchmarks from receiving
greater weight solely because of their size.

\subsection{Shared task results}
\label{sec:shared_task_results}

Table~\ref{tab:shared_task_results} reports the official team-deduplicated final
leaderboard.  The best award-eligible score in each aggregate column is shown in bold.

\begin{table}[htbp]
\centering
\footnotesize
\begingroup
\setlength{\tabcolsep}{3pt}
\renewcommand{\arraystretch}{1.04}
\begin{CJK*}{UTF8}{gbsn}
\begin{tabularx}{\textwidth}{
    @{}
    r
    >{\raggedright\arraybackslash}p{0.18\textwidth}
    >{\raggedright\arraybackslash}X
    l
    rrrr
    @{}
}
\toprule
Rank & Team & Model & Arch. & NLU & Cog & Hanzi & Overall \\
\midrule
1 & BabyDragon & BabyDragon-v4 & DeBERTa-v2 &
59.98 & 32.94 & \textbf{91.56} & \textbf{61.49} \\
2 & OK的老大 & babydeberta-24m-hanzi-v8s & BERT &
58.55 & 33.35 & 81.46 & 57.79 \\
3 & TUM Campus Heilbronn & babylm-zho-pinyin-code-97M & GPT-2 &
56.40 & 32.80 & 75.51 & 54.90 \\
4 & SMI & Chinese\_babylm & BERT &
60.13 & 33.78 & 61.41 & 51.77 \\
5 & lampa & chinese-babylm-2026-hanzi-a3large & GPT-2 &
59.09 & 33.32 & 58.70 & 50.37 \\
6 & YNU-HPCC & BabyLM-Chinese-DeBERTa-V2-14M & DeBERTa-v2 &
\textbf{60.99} & 33.29 & 55.71 & 50.00 \\
7 & 西湖往事 & babylm-zh-run10a & GPT-2 &
60.75 & 33.27 & 54.69 & 49.57 \\
8 & L-Cluster & chinese-babylm-2026-v3 & BERT &
58.65 & 33.11 & 56.49 & 49.42 \\
9 & moddells & moddells/babylm & BERT &
58.99 & 33.67 & 52.70 & 48.45 \\
10 & CogTeam & babyconceptLM-baike102M & BERT &
56.79 & 33.76 & 53.61 & 48.05 \\
11 & Alice & Chinese BabyLM NanoChat D22 & GPT-2 &
56.87 & 33.97\footnote{This team's submitted result used the best layer.} & 50.38 & 47.07 \\
12 & Flops & pinyinboost-bert-10M-6L384H-stage6e-vocabadapt & BERT &
52.34 & 32.16 & 51.59 & 45.36 \\
13 & KABUTOMUSHI & chinese-babylm-bpe-11 & BERT &
53.21 & 30.38 & 49.96 & 44.52 \\
14 & 所以然 & syr\_model\_bert\_final\_3 & BERT &
50.14 & 30.91 & 50.10 & 43.72 \\
15 & XJTLU-SeedLM & GPT2-ChineseDevBench & GPT-2 &
43.98 & 34.10\footnote{This submission was evaluated only on open tasks and data and was therefore ineligible for an award.} & 27.67 & 35.25 \\
16 & ZZUnlp & public\_my\_babys10 & BERT &
59.02 & 33.35 & 1.19 & 31.19 \\
17 & Deepscientist & chinese-babylm-cog-a197-strict-best & BERT &
0.00 & \textbf{33.96} & 0.00 & 11.32 \\
18 & kiikii & neurobert-cog-v2 & BERT &
0.00 & 33.00 & 0.00 & 11.00 \\
\bottomrule
\end{tabularx}
\end{CJK*}
\endgroup
\caption{Official results of the ChineseBabyLM shared task. NLU, Cog, and
Hanzi are the mean scores for the three evaluation tracks; Overall is their
macro-average. The best score in each aggregate column is shown in bold.}
\label{tab:shared_task_results}
\end{table}

According to the submitted system metadata and technical report, BERT is the most common
architecture, accounting for 11 of the 18 ranked systems.  Five systems use
GPT-2 and two use DeBERTa-v2; grouping BERT and DeBERTa-v2 together, BERT-family
architectures therefore account for 13 of the 18 entries.  

The two highest-ranked systems are encoder-only, whereas the third-ranked system is decoder-only. All three track-winning systems are encoder-only, although the strong overall performance of the third-ranked decoder-only system shows that competitive submissions were not limited to a single model type.


The top submissions to the first ChineseBabyLM challenge illustrate three complementary approaches to learning Chinese under a limited-data regime: writing-system-aware pretraining, brain-aligned representation synthesis, and compact curriculum-trained language modeling.

\textbf{BabyDragon-v4}, submitted by the BabyDragon team, ranked first overall
and first on the Hanzi track~\citep{zhao2026babydragon}.  It is a 104M-parameter
DeBERTa-v2 encoder trained from scratch for 60 epochs with whole-word masked
language modeling and an auxiliary pinyin-prediction objective.  Its 92.91M-word
training mixture contains 6.76M words of injected facts.  Pinyin and character
structure facts were derived from resources such as \texttt{pypinyin},
CHISE/CJKVI-IDS, and Unihan, while grammar examples were generated by rule; no
large language model was used to generate or paraphrase the data.  Controlled
analysis indicates that the model
generalizes well to homophone pairs not explicitly injected, whereas its
orthographic decomposition performance depends much more strongly on coverage
of the injected character facts.  Although the tokenizer was extended with 329
radical and component tokens, this extension was effectively inactive on the
final evaluation because all tested components were already covered.

\textbf{chinese-babylm-cog-a197-strict-best} (RepDuet)~\citep{liu2026repduet}
won the Cognitive Modeling track after organizer reproduction and
final-layer verification.  The two higher raw Cog entries
were excluded from the award because one used only open tasks and data and the
other did not submit final-layer representations.  RepDuet uses two compact
BERT encoders trained from scratch on the official corpus and constructs a
different representation for each fMRI task.  The word-level representation
concatenates intermediate contextual states with random-indexing and token
frequency features.  The discourse-level representation combines an
upper-layer state mixture with lexical statistics and MLM-derived confidence
features, including token likelihood, entropy, maximum probability, and
prediction margin.  This task-specific design achieved a verified Cog score of
33.96.

\textbf{BabyLM-Chinese-DeBERTa-V2-14M} (BabyDeBERTa-zh)~\citep{lu2026babydeberta}
won the NLU track with a score of 60.99.  It is a
14M-parameter DeBERTa-v2 masked language model with 12 layers, a hidden size of
256, a feed-forward size of 1024, and 8 attention heads.  Its 12,800-token
WordPiece tokenizer was retrained on the official corpus.  The main innovation
is a deterministic four-stage curriculum that restricts complete training
sequences to maximum lengths of 32, 64, and 128 before exposing the model to
the full-length corpus.  In the final 30-epoch run, the four stages receive 2,
3, 5, and 20 epochs, respectively.  Compared with an otherwise comparable
non-curriculum baseline, this schedule raises the NLU average from 59.70 to
60.99, with the largest improvement on C3.

\textbf{babydeberta-24m-hanzi-v8s} (KADeBERTa-v2)~\citep{cai2026kadeberta}
ranked second overall and second on the Hanzi track.  It is a compact DeBERTa-v2 encoder with
approximately 26M parameters, 10 layers, a hidden size of 384, and 6 attention
heads.  Its custom 12,119-token WordPiece vocabulary preserves Chinese
character components as atomic units.  The model augments the official corpus
with automatically generated pinyin and character-structure statements and
uses weighted sampling to expose the model more frequently to this small
auxiliary corpus during masked-language-model pretraining.  Ablation results
show that phonological facts primarily improve pinyin judgments, whereas
structural facts primarily improve character-structure judgments.

\textbf{babylm-zho-pinyin-code-97M}, submitted by TUM Campus Heilbronn,
ranked third overall~\citep{robrecht2026pinyin}.  The system replaces Hanzi input with an \emph{Initial+Digit}
representation that maps each pinyin syllable to two ASCII symbols encoding its
initial and coarse information about tone and syllable length.  Codes within
each Jieba-segmented word are concatenated, while word boundaries are retained;
the same deterministic conversion is applied to evaluation inputs.  A
16,000-token SentencePiece BPE tokenizer is trained on this representation and
used with a 97.7M-parameter, 12-layer decoder-only GPT model optimized by
causal language modeling.  Its results show that a heavily compressed phonetic
representation can retain enough information for competitive Chinese language
modeling despite discarding Hanzi identity and most orthographic information.

\section{Discussion}
\label{sec:discussion}

\subsection{Approaches of the top systems}
\label{subsec:discussion_models}

The five award-winning systems adopt complementary strategies.  The two
highest-ranked Hanzi systems, BabyDragon-v4 and KADeBERTa-v2, both improve a
standard DeBERTa-v2 encoder primarily through \emph{data augmentation},
injecting rule-generated pinyin and character-structure facts into pretraining;
their success confirms that the character-level knowledge probed by the Hanzi
track is largely absent from ordinary child-directed text and must be supplied
explicitly.  The NLU winner, BabyDeBERTa-zh, takes a different route, relying on
a sequence-length curriculum rather than augmented data.  The Cognitive Modeling
winner, RepDuet, changes neither the model nor the data but the
\emph{representations used at evaluation}, combining several compact BERT
checkpoints into task-specific feature sets.  The third-ranked system,
babylm-zho-pinyin-code-97M, is the only top system to innovate on the
\emph{tokenizer and input representation} itself, replacing Hanzi with a
compressed phonetic code that retains enough information for competitive
language modeling while discarding orthographic identity.

This contrast points to several directions for future modeling.  Because the
Hanzi winners relied on data augmentation, an open question is whether comparable
gains can be obtained through architectural or objective-level inductive biases
for phonology and orthography rather than injected facts.  More broadly,
babylm-zho-pinyin-code-97M shows that representation design is a promising but
underexplored axis: future systems could explore tokenizers that encode
orthographic, phonological, and semantic information natively, potentially
without external augmentation.  Finally, as each award-winning system was
optimized for a single track, building one model that is competitive across NLU,
Hanzi, and cognitive modeling remains an open goal.

\subsection{Task discriminability}
\label{subsec:discussion_tasks}

The three tracks differ markedly in how well they separate systems.  Across the
18 ranked submissions, the Hanzi track spans roughly 90 points (1.19--91.56) and
the NLU track 17 points (43.98--60.99 among the teams that submitted it),
whereas the Cognitive Modeling track spans fewer than 4 points (30.38--34.10,
standard deviation below 1).  The Cognitive Modeling compression is driven by its
constituent tasks: the word-level fMRI task varies by only 1.2 points over the
entire field (55.47--56.65), and the discourse-level task, although somewhat
wider (5.17--11.90), does so at low absolute scores.  In its current form the
track therefore contributes little to the overall ranking once a system produces
reasonable representations.

Several NLU tasks are similarly weak at discriminating systems.  AFQMC varies by
only 2.3 points across all submitting teams (68.86--71.11), with most systems
within a single point of one another---the same label-bias degeneracy observed
in the baselines, now seen across the whole field, indicating that this task
measures little about the models.  CLUEWSC2020 is also compressed
(62.83--68.75).  By contrast, the minimal-pair and inference tasks discriminate
well: ZhoBLiMP spans 29 points (53.21--82.21), OCNLI 19 (53.56--72.68), and the
diagnostic NLI task 16 (38.31--54.48).  The Hanzi tasks separate the field
strongly but exhibit ceiling effects among the top systems, with two teams
reaching 99.85 and 95.80 on open pinyin.

These observations motivate a revision of the task inventory in future editions.
Tasks that fail to discriminate among systems---\textit{word\_fmri}, AFQMC, and
CLUEWSC2020 being the clearest cases---should be revised or retired, while
high-spread tasks such as ZhoBLiMP, the NLI tasks, and the Hanzi tasks should be
retained and, where they saturate, made harder.  Adding new tasks that target
underrepresented Chinese phenomena, such as word segmentation, morphology, or
discourse coherence, or developmental tasks such as Age of Acquisition prediction, 
would further increase the diagnostic value of the
benchmark.

\subsection{Comparison with the English BabyLM Challenge}
\label{subsec:discussion_english}

Comparing the first ChineseBabyLM Challenge with the English BabyLM shared
tasks highlights both shared and language-specific design choices.

\paragraph{Model architectures.}
The dominant architectures differ in instructive ways.  In the English BabyLM
challenges, decoder-only causal models and hybrid objectives were prominent
among the leading systems; in particular, GPT-BERT, which combines causal and
masked language modeling in a single stack, was a winning
entry~\citep{charpentier2024gptbert}.  In ChineseBabyLM, by contrast, no team
adopted such a causal/bidirectional hybrid: encoder-only BERT and DeBERTa-v2
systems dominated, accounting for 13 of the 18 ranked entries.  This divergence
is consistent with a controlled finding from the top-ranked BabyDragon team,
who reported that adding a GPT-BERT-style causal branch was the weakest of their
architectural variants, lowering rather than raising performance~\citep{zhao2026babydragon}.  They
attribute this to the all-bidirectional nature of the ChineseBabyLM evaluation,
in which a causal signal dilutes the masked objective without an evaluation
payoff.  The contrast suggests that the value of a modeling innovation depends
on the evaluation regime, and that recipes validated on English may not transfer
automatically to Chinese.

\paragraph{Task selection.}
The task inventories overlap but diverge where the languages do.  Both
challenges include a natural-language-understanding track built on grammatical
minimal pairs (BLiMP for English, ZhoBLiMP for Chinese) and supervised
classification, and both evaluate cognitive plausibility.  Beyond this shared
core, the English challenge additionally includes behavioral psycholinguistic
tasks such as age-of-acquisition prediction and reading-time
modeling~\citep{warstadt-etal-2023-findings,hu-etal-2024-findings,charpentier-etal-2025-findings},
which ChineseBabyLM does not yet have.  Conversely, the Hanzi track is specific
to Chinese, probing orthographic structure and phonology that have no direct
English counterpart.  Future ChineseBabyLM editions could therefore both expand
the shared cognitive component---for example by adding Chinese reading-time or
acquisition-trajectory tasks to parallel the English setup---and deepen the
Chinese-specific component that currently distinguishes the two benchmarks.

\section{Conclusion and future directions}
In this paper, we have described the design and results of the first
ChineseBabyLM Challenge, which was created to promote data-efficient and
cognitively plausible language modeling for Chinese.  The submitted systems
illustrate several complementary strategies for learning under a limited-data
regime: compact DeBERTa-style encoders; Chinese-specific tokenization,
pinyin-aware objectives, and explicit character-knowledge augmentation;
curriculum learning; compressed phonetic input representations; and
task-specific combinations of contextual, lexical, and uncertainty features
for cognitive modeling.

These findings suggest several directions for future ChineseBabyLM research.
First, systematic ablations would help clarify the contribution of knowledge
injection, tokenizer design, pinyin-aware objectives, curriculum schedules,
phonetic recoding, and multi-checkpoint representation fusion.  Second, future
challenges could encourage unified models that perform well across NLU, Hanzi,
and cognitive modeling rather than optimizing for a single track.  Finally,
Chinese offers rich opportunities for language-specific evaluation beyond
standard sentence understanding, including orthographic structure, phonology,
morphology, reading difficulty, and developmental acquisition.  Incorporating
these factors into both pretraining and evaluation may produce models that are
not only more accurate but also more informative as computational models of
Chinese language learning.

\setlength{\bibsep}{0pt plus 0.3ex}%
\renewcommand{\bibfont}{\small}%
\bibliography{main-short}

\end{document}